# PROBABILITY JUDGMENT IN ARTIFICIAL INTELLIGENCE

Glenn Shafer
School of Business, University of Kansas

This paper is concerned with two theories of probability judgment: the Bayesian theory and the theory of belief functions. It illustrates these theories with some simple examples and discusses some of the issues that arise when we try to implement them in expert systems.

The Bayesian theory is well known; its main ideas go back to the work of Thomas Bayes (1702-1761). The theory of belief functions, often called the Dempster-Shafer theory in the artificial intelligence community, is less well known, but it has even older antecedents; belief-function arguments appear in the work of George Hooper (1640-1723) and James Bernoulli (1654-1705). For elementary expositions of the theory of belief functions, see Shafer (1976,1985).

## 1. Two Probability Languages.

The Bayesian theory and the theory of belief functions both use the standard calculus of probability, but they apply this calculus to concrete problems in different ways. The difference is best understood in the context of the general philosophy of constructive probability (Shafer, 1981; Shafer and Tversky, 1985).

According to the philosophy of constructive probability, numerical probability judgment involves fitting a problem to a scale of canonical examples. The canonical examples usually involve the picture of chance in some way, but different choices of canonical examples are possible, and these different choices produce different theories or, if you will, different languages in which to express probability judgments. No matter what language is used, the judgments are subjective; the subjectivity enters when we judge that the evidence in our actual problem matches in strength and significance the evidence in the canonical example.

In the Bayesian probability language, the canonical examples are examples in which the answer to the question with which we are concerned is determined by chance, and the chances of the different possible answers are known. Within this language, many different designs for probability judgment are possible. Some designs, which we may call <u>total-evidence designs</u>, fit our evidence about a question to these canonical examples by comparing it directly to knowledge that the answer is determined by known chances. Other designs, which we may call <u>conditioning designs</u>, initially compare only part of our evidence to such knowledge and then take the remaining evidence into account by conditioning.

Much that is written about Bayesian methods pretends that we come to practical problems knowing which part of our evidence is to be taken into account by conditioning. Sometimes it is even suggested that all of our evidence should be taken into account by conditioning. But in fact the choice of which evidence to take into account by conditioning and which to use in assessing initial probabilities is usually the most important choice in the design of a Bayesian analysis.

It may be useful to elaborate this point. Suppose we want to make probability judgments about a frame of discernment S. (A <u>frame of discernment</u> is a list of possible answers to a question; so this means we want to make probability judgments about which answer is correct.) We reflect on what relevant evidence we have, and produce a list $E_1,\ldots,E_n$ of facts that seem to summarize this evidence adequately. A conditioning design might ask us to stand back from our knowledge of these n facts, pretend that we do not yet know them, and construct a probability measure over a frame that considers not only the question considered by S but also the question whether $E_1,\ldots,E_n$ are or are not true; often we construct this measure by making probability judgments $P(s)$ and $P(E_1\&\ldots\&E_n|s)$ for each s in S. The problem with this strategy is that we now need to look for evidence on which to base these probability judgments. We have used our best evidence up, as it were, but now we have an even larger judgmental task than before. According to some theorists, there is no problem-- it is normative to have the requisite probabilities, whether we can identify relevant evidence or not. But according to the constructive philosophy, there is a problem, a problem which limits how far we want to go. We may want to condition only on some of the $E_i$, reserving the others to help us make the probability judgments.

Whereas the Bayesian probability language uses canonical examples where known chances are attached directly the possible answers to the question asked, the language of belief functions uses canonical examples where known chances may be attached only to the possible answers to a related question.

Suppose, indeed, that S and T denote, respectively, the possible answers to two distinct but related questions. When we say that these questions are related, we mean that a given answer to one of the questions may not be compatible with all the possible answers to the other. Let us write "sCt" when s is an element of S, t is an element of T, and s and t are compatible. Given a probability measure P over S (we assume for simplicity that P is defined for all subsets of S), we may define a function Bel on subsets of T by setting

$$Bel(B) = P\{s | \text{if } sCt, \text{ then } t \text{ is in } B\}. \qquad (1)$$

for each subset B of T. The right-hand side of (1) is the probability that P gives to those answers to the question considered by S that require the answer to the question considered by T to be in B; the idea behind (1) is that this probability should be counted as reason to believe that the latter answer is in B. We might, of course, have more direct evidence about the question considered by T, but if we do not, or if we want to leave other evidence aside for the moment, then we may call Bel(B) a measure of the reason we have to believe B based just on P.

We call the function Bel given by (1) the <u>belief function</u> obtained by extending P from S to T. A probability measure P is a special kind of belief function; this is just the case where ① S=T and ② sCt if and only if s=t.

All the usual devices of probability are available to the language of belief functions, but in general they are applied in the background, at the level of S, before extending to degrees of belief on T, the frame of interest. Thus the language of belief functions is a generalization of the Bayesian language.

## 2. Three Examples.

<u>Example 1</u>. Is Fred, who is about to speak to me, going to speak truthfully, or is he, as he sometimes does, going to speak carelessly, saying something that comes into his mind, paying no attention to whether it is true or not? Let S denote the possible answers to this question; S={truthful,careless}. Suppose I know from experience that Fred's announcements are truthful reports on what he knows about 80% of the time and are careless statements the other 20% of the time. Then I have a probability measure P over S: P{truthful}=.8, P{careless}=.2.

Are the streets outside slippery? Let T denote the possible answers to this question; T={yes,no}. And suppose Fred's announcement turns out to be, "The streets outside are slippery." Taking account of this, I have a compatibility relation between S and T; "truthful" is compatible with "yes" but not with "no," while "careless" is compatible with both "yes" and "no." Applying (1), I find

$$Bel(\{yes\})=.8 \quad \text{and} \quad Bel(\{no\})=0; \qquad (2)$$

Fred's announcement gives me an 80% reason to believe the streets are slippery outside, but no reason to believe they are not.

How might a Bayesian argument using this evidence go? A total-evidence design would use all my evidence, Fred's announcement included, to make a direct probability judgment about whether the streets are slippery. But if I want an argument that uses the judgment that Fred is 80% reliable as one ingredient, then I will use a conditioning design that requires two further probability judgments: (1) A prior probability, say p, for the proposition that the streets are slippery; this will be a judgment based on evidence other than Fred's announcement. (2) A conditional probability, say q, that Fred's announcement will be accurate even though it is careless. Given these ingredients, I can calculate a Bayesian probability that the streets are slippery given Fred's announcement and my other evidence:

$$P(\text{slippery}|\text{announcement}) = \frac{.8p + .2pq}{.8p + .2pq + .2(1-p)(1-q)}. \quad (3)$$

Is the Bayesian argument (3) better than the belief-function argument (2)? This depends on whether I have the evidence required. If I do have evidence to support the judgments p and q--if, that is to say, my situation really is quite like a situation where the streets and Fred are governed by known chances, then (3) is a good argument, clearly more convincing than (2) because it takes more evidence into account. But if the evidence on which I base p and q is of much lower quality than the evidence on which I base the number 80%, then (2) will be more convincing.

The traditional debate between the frequentist and Bayesian views has centered on the quality of evidence for prior probabilities. It is worth remarking, therefore, that we might well feel that q, rather than p, is the weak point in the argument (3). I probably will have some other evidence about whether it is slippery outside, but I may not have any idea about how likely it is that Fred's careless remarks will accidentally be true.

A critic of the belief-function argument (2) might be tempted to claim that the Bayesian argument (3) shows (2) to be wrong even if I do lack the evidence needed to supply p and q. Formula (3) gives the correct probability for whether the street is slippery, the critic might contend, even if I cannot say what this probability is, and it is almost certain to differ from (2). This criticism is fundamentally misguided. In order to say that (3) gives the "correct" probability, I must be able to convincingly compare my situation to the picture of chance. And my inability to model Fred when he is being careless is not just a matter of not knowing the chances--it is a matter of not being able to fit him into a chance picture at all.

Example 2. Suppose I do have some other evidence about whether the streets are slippery: my trusty indoor-outdoor thermometer says that the temperature is 31 degrees Fahrenheit, and I know that because of the traffic, ice could not form on the streets at this temperature.

My thermometer could be wrong. It has been very accurate in the past, but such devices do not last forever. Suppose I judge that there is a 99% chance that the thermometer is working properly, and I also judge that Fred's behavior is independent of whether it is working properly or not. (For one thing, he has not been close enough to my desk this morning to see it.) Then I have determined probabilities for the four possible answers to the question, "Is Fred being truthful or careless, and is the

94

thermometer working properly or not?" For example, I have determined the probability .8x.99=.792 for the answer "Fred is being truthful, and the thermometer is working properly." All four possible answers, together with their probabilities, are shown in the first two columns of Table 1. We may call the set of these four answers our new frame S.

Taking into account what Fred and the thermometer have said, I have the compatibility relation between S and T given in the last column of the table. (Recall that T considers whether the streets are slippery; T={yes,no}.) The element (truthful,working) of S is ruled out by this compatibility relation (since Fred and the thermometer are contradicting each other, they cannot both be on the level); hence I condition the initial probabilities by eliminating the probability for (truthful,working) and renormalizing the three others. The resulting posterior probabilities on S are given in the third column of the table.

Finally, applying (1) with these posterior probabilities on S, I obtain the degrees of belief

$$Bel(\{yes\})=.04 \quad \text{and} \quad Bel(\{no\})=.95. \qquad (4)$$

This result reflects that fact that I put much more trust in the thermometer than in Fred.

The preceding calculation is an example of <u>Dempster's rule of combination</u> for belief functions. Dempster's rule combines two or more belief functions defined on the same frame but based on independent arguments or items of evidence; the result is a belief function based on the pooled evidence. In this case the belief function given by (2), which is based on Fred's testimony alone, is being combined with the belief function given by

$$Bel(\{yes\})=0 \quad \text{and} \quad Bel(\{no\})=.99, \qquad (5)$$

which is based on the evidence of the thermometer alone. In general, as in this example, Dempster's rule corresponds to the formation and subsequent conditioning of a product measure in the background. See Shafer (1985) for a precise account of the independence conditions needed for Dempster's rule.

| s | Probability of s | | Elements of T compatible with s |
|---|---|---|---|
| | Initial | Posterior | |
| (truthful,working) | .792 | 0 | -- |
| (truthful,not) | .008 | .04 | yes |
| (careless,working) | .198 | .95 | no |
| (careless,not) | .002 | .01 | yes,no |

Table 1.

Example 3. Dempster's rule applies only when two items of evidence are independent, but belief functions can also be derived from models for dependent evidence.

Suppose, for example, that I do not judge Fred's testimony to be independent of the evidence provided by the thermometer. I exclude the possibility that Fred has tampered with the thermometer and also the possibility that there are factors affecting both Fred's truthfulness and the thermometer's accuracy. But suppose now that Fred does have regular access to the thermometer, and I think that he would likely know if it were not working. I know from experience that it just in situations like this, where something is awry, that Fred tends to let his fancy run free.

In this case, I would not assign the elements of S the probabilities given in the second column of Table 1. Instead, I might assign the probabilities given in the second column of Table 2. These probabilities follow from my judgment that Fred is truthful 80% of the time and that the thermometer has a 99% chance of working, together with the further judgment that Fred has a 90% chance of being careless if the thermometer is not working.

When I apply (1) with the posterior probabilities given in Table 2, I obtain the degrees of belief

$$Bel(\{yes\})=.005 \quad \text{and} \quad Bel(\{no\})=.95.$$

These differ from (4), even though the belief functions based on the separate items of evidence are still be given by (2) and (5).

I would like to emphasize that nothing in the philosophy of constructive probability or the language of belief functions requires us to deny the fact that Baysian arguments are often valuable and convincing. The examples I have just discussed were designed to convince the reader that belief-function arguments are sometimes more convincing than Bayesian arguments, not that this is always or even usually the case. What the language of belief functions does require us to reject is the philosophy according to which use of the Bayesian language is normative.

| s | Probability of s | | Elements of T compatible with s |
|---|---|---|---|
| | Initial | Posterior | |
| (truthful,working) | .799 | 0 | -- |
| (truthful,not) | .001 | .005 | yes |
| (careless,working) | .191 | .950 | no |
| (careless,not) | .009 | .045 | yes,no |

Table 2.

From a technical point of view, the language of belief functions is a generalization of the Bayesian language. But as our examples illustrate, the spirit of the language of belief functions can be distinguished from the spirit of the Bayesian language by saying that a belief-function argument involves a probability model for the evidence bearing on a question, while a Bayesian argument involves a probability model for the answer to the question.

Of course, the Bayesian language can also model evidence. As we have seen in our examples, the probability judgments made in a belief-function argument can usually be adapted to a Bayesian argument that models both the answer to the question and the evidence for it by assessing prior probabilities for the answer and conditional probabilities for the evidence given the answer. The only problem is that we may lack the evidence needed to make all the judgments required by this Bayesian argument convincing. Thus we may say that the advantage gained by the belief-function generalization of the Bayesian language is the ability to use certain kinds of incomplete probability models.

### 3. Probability Judgment in Expert Systems.

In statistics and many other fields, it is often taken for granted that "subjective probability" means Bayesian probability. But most of the expert systems developed during the past ten years have not used full-fledged Bayesian designs, and the creators of these systems have shown great interest in belief functions and other alternatives to Bayesian methods. Why is this?

We can gain some insight into this question by noting that the idea of an expert system developed from efforts to apply systems of production rules to practical problems. Workers in artificial intelligence were attracted to production systems, or systems of if-then rules, because the modular character of these rules and the unstructured programming that they allowed seemed to correspond to the flexibility in acquiring and using knowledge that is characteristic of human intelligence.

It is clear that a Bayesian design does not have this modular character. We are not free to add or remove probability judgments from a Bayesian design in the way that we are free to add or remove production rules from a production system. A Bayesian design specifies very rigidly just what probability judgments it requires.

The designers of early systems such as PROSPECTOR and MYCIN were forced to compromise between modularity and conformity with the standards of Bayesian design. In the case of PROSPECTOR, the result was a set of quasi-Bayesian rules for combining probabilities, together with some ad hoc restrictions on the modularity of the system. In the case of MYCIN, the result was an ad hoc set of rules for combining "certainty factors," rules that turned out to be very similar to the rules for belief functions (Gordon and

Shortliffe, 1984).

The interest in belief functions among designers of expert systems can be attributed in large measure to the greater modularity it permits. Dempster's rule allows us to formally add or remove any item of evidence in a belief-function analysis, provided only that it is independent of the other items of evidence being combined.

The philosophy of constructive probability makes it clear, however, that this greater modularity is only a matter of degree. All probability judgment, whether Bayesian or belief-function, requires a design. In the Bayesian case we must decide what to condition on. In both cases we must make either make judgments of independence or else construct more complicated designs that take dependence into account.

I believe that in the next few years both Bayesian and belief-function designs will find their niches in the world of expert systems. Bayesian designs will predominate in systems that are repeatedly applied under conditions so constant that the picture of answers determined at random with known chances fits. Belief-function designs will be more successful in systems whose each use represents a relatively unusual conjunction of different small worlds of experience. All these systems will, however, fit only certain narrow kinds of problems. They will fall far short of the flexibility of human probability judgment.

Human probability judgment is flexible not because it is modular, but because it includes the capacity to design probability arguments as well as the capacity to carry out such designs. The deepest and most intriguing challenge for probability in artificial intelligence is the development of this capacity for design.